\documentclass[11pt]{article}
\usepackage[final]{acl}
\usepackage{times}
\usepackage{latexsym}
\usepackage{graphicx}
\usepackage{booktabs}
\usepackage{amsmath}
\usepackage{multirow}
\usepackage{url}
\usepackage{xcolor}
\usepackage{fontawesome5}
\usepackage{microtype}
\usepackage{adjustbox}
\newcounter{algorithm}
\newcommand{\algcaption}[1]{\refstepcounter{algorithm}\noindent\textbf{Algorithm~\thealgorithm: #1}}
\newcommand{\equalmark}{%
  \textsuperscript{\scriptsize\faHandshake}%
}
\newcommand{\corrmark}{%
  \textsuperscript{\scriptsize\faIcon{envelope}}%
}

\title{When the Weakest Language Keeps Moving: ARMOR-DP for Multilingual BabyLM}

\author{
Hanyu Liao$^2$\equalmark
\quad
Yijing Chen$^1$\equalmark
\quad
Tao Ji$^1$\corrmark
\\
$^1$Fudan University \\
$^2$East China Normal University
\\
\small{
    \href{mailto:51270400017@stu.ecnu.edu.cn}{51270400017@stu.ecnu.edu.cn} \quad\quad
    \href{mailto:yijingchen24@m.fudan.edu.cn, taoji@fudan.edu.cn}{\{yijingchen24@m., taoji@\}fudan.edu.cn}
   }
}

\begin{document}
\maketitle

\begingroup
\renewcommand{\thefootnote}{\faHandshake}
\footnotetext{Equal contribution.}
\endgroup

\begingroup
\renewcommand{\thefootnote}{\faIcon{envelope}}
\footnotetext{Corresponding author.}
\endgroup

\setcounter{footnote}{0}

\begin{abstract}
Multilingual BabyLM systems usually fix language sampling proportions even
though the weakest language can change during training. We present ARMOR-DP, a
modular framework that diagnoses imbalance arising from representation,
exposure, capacity, and objective pressure, and uses per-language validation
loss to update sampling. Controlled experiments show that equal character
budgets create a severe representation artifact for Chinese; byte-level
encoding removes much of it, while static Chinese protection can subsequently
make English the weakest language. The dynamic controller tracks these
transitions and achieves a better score--balance tradeoff than fixed
protection. On official evaluation, curriculum-dynamic checkpoints outperform
a matched static Chinese-protective baseline while using fewer Chinese epochs;
scaling capacity yields the strongest internal ARMOR-DP system. These results
support a simple conclusion: multilingual weakness is a training-time state
rather than a fixed property of a language. ARMOR-DP therefore separates
diagnosis from intervention, allowing protection to follow the current
bottleneck instead of a predetermined language label.
\end{abstract}

\section{Introduction}

The BabyLM challenge tasks a model with acquiring language from a
developmentally plausible data budget, and the 2026 BabyBabelLM track extends
this premise to the multilingual setting
\citep{babylm2026guidelines,babybabellm2026}. Naturalistic linguistic input is
not a fixed curriculum: properties of child-directed speech change over
development and can track children's emerging linguistic abilities
\citep{vosoughi2012prosody}. We use this responsiveness as a motivating analogy,
not as a claim that caregivers optimize explicit language proportions.

\begin{figure}[t]
  \centering
  \makebox[\linewidth][c]{\includegraphics[width=1\linewidth]{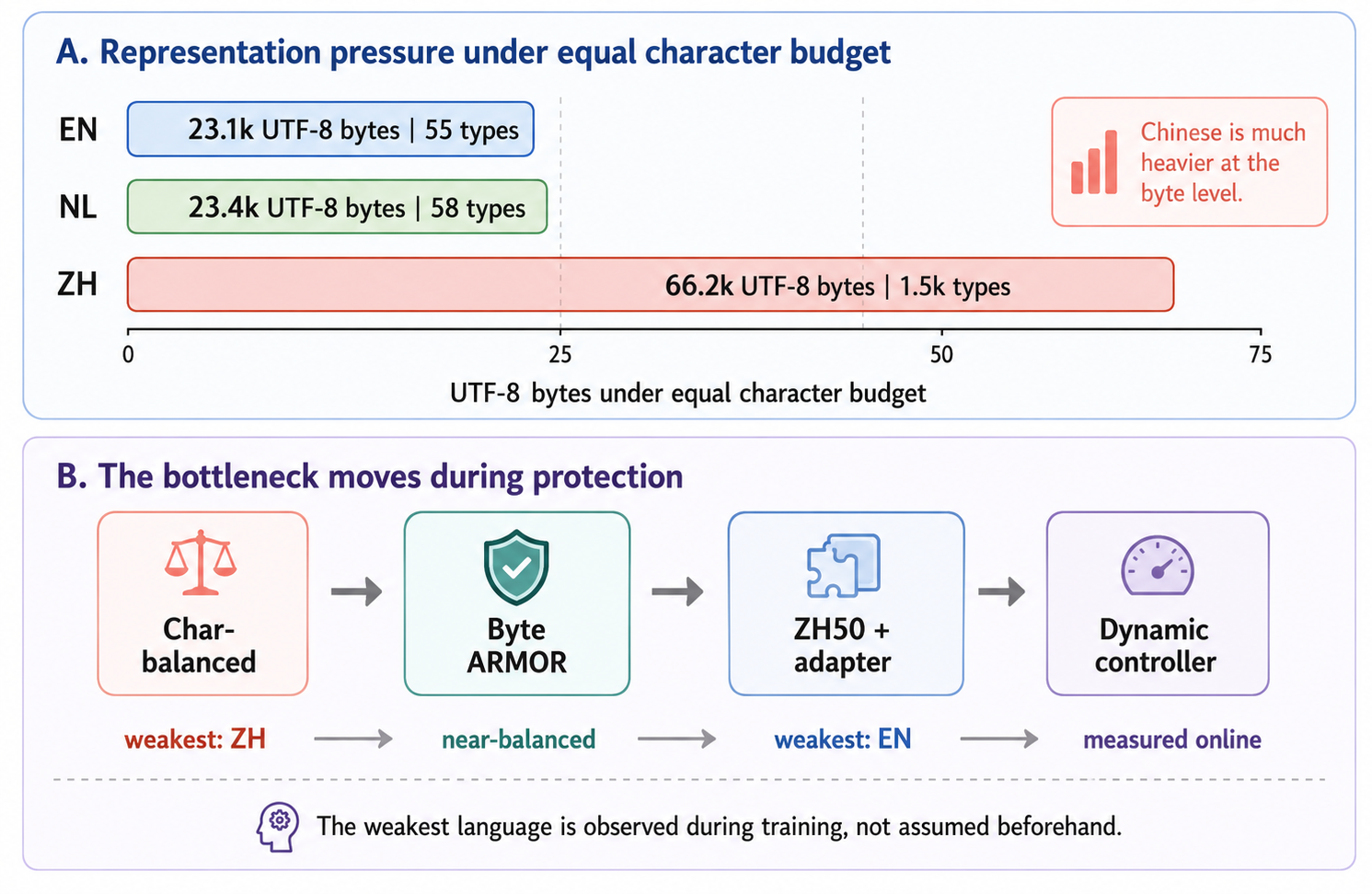}}
  \caption{Central hypothesis of ARMOR-DP. Panel A shows that equal character
  budgets do not create equal representation burden across scripts; Panel B
  shows that the bottleneck can move after protection is added. The takeaway is
  that weakest-language protection should be measured during training rather
  than fixed to one language in advance.}
  \label{fig:overview}
\end{figure}

In contrast, standard multilingual pretraining relies on static sampling
temperatures or predetermined corpus proportions, which are agnostic to the
model's evolving proficiency. Under BabyLM's stringent capacity and data
constraints, languages compete for a shared parameter budget, and a fixed ratio
can underserve whichever language currently lags. A responsive artificial
learner should therefore reallocate exposure online. Doing so requires first
diagnosing why a language is weak at a particular moment---a question that
cannot be answered by exposure alone.

Our central hypothesis is that a language can be \textit{weak} for at least four different reasons. It can be weak because its script is poorly represented by the tokenizer (\textbf{representation}), because it receives insufficient effective \textbf{exposure}, because the shared model has insufficient language-private \textbf{capacity}, or because the objective allocates gradient pressure in a way that amplifies already frequent units (\textbf{objective}). Therefore, we propose \textbf{ARMOR-DP} (\textbf{Armor} for \textbf{D}evelopmental \textbf{P}ressure), a modular diagnostic and intervention framework.

Our contributions are threefold. First, we show that the identity of the
weakest language moves as representation, exposure, and capacity interventions
take effect. Second, ARMOR-DP separates four possible sources of weakness and
turns per-language validation loss into an online sampling signal. Third, our
official evaluations show that curriculum-dynamic training improves on a
matched static Chinese-protective system while remaining within the challenge
budget.

\section{ARMOR-DP Framework}

ARMOR-DP is a modular diagnostic and intervention toolkit. It separates four
sources of multilingual weakness, each paired with a corresponding probe or
intervention. Components can be enabled or omitted according to empirical
evidence rather than treated as a mandatory stack.

\paragraph{Representation armor.}
We use UTF-8 byte units as a script-fair representation. It removes a key comparison artifact, in which Chinese characters are treated as sparse atomic symbols while
Latin-script languages reuse a small alphabet. This choice is based on 
tokenization-free and byte-level models \citep{xue2022byt5,clark2022canine,tay2022charformer}.

\paragraph{Exposure armor.}
We compare balanced sampling with ZH-protective schedules. Our primary static baseline uses English/Dutch/Chinese ratios of 25/25/50. We also test a
curriculum that starts balanced and switches to 25/25/50 halfway through
training. This is related to temperature-based multilingual sampling and
curriculum learning \citep{conneau2020xlmr,bengio2009curriculum}.

\paragraph{Capacity armor.}
We test two lightweight capacity mechanisms: language adapters and a
language-conditioned routing gate over attention heads. We employ the routing gate as an interpretable capacity intervention for multilingual adaptation. Rather than introducing a new routing mechanism, our goal is to analyze how capacity allocation interacts with language imbalance within the ARMOR-DP framework \citep{houlsby2019adapters,pfeiffer2020madx,fedus2022switch}.

\paragraph{Objective armor.}
We test both a naive rare-token weighted objective and a pressure-aware
objective that combines current excess loss, script pressure, and exposure. The
rare-token diagnostic multiplies token-level cross-entropy by a weight derived
only from that byte type's frequency in the diagnostic corpus: less frequent
types receive larger weights, clipped to the range reported in
Appendix~\ref{tab:hyperparams}. It is included as a simple frequency-only
analogue of loss reweighting \citep{lin2017focal}, not as part of the final
checkpoint. In our diagnostics it over-protects Chinese after exposure has
already helped it, whereas pressure-aware weighting reduces this failure.

\paragraph{Dynamic protection controller.}
The dynamic controller is responsible for adapting the sampling distribution as multilingual learning progresses. Instead of assuming that the weakest language remains fixed throughout training, it continuously estimates the current learning pressure of each language and reallocates training probability accordingly. At each update step \(t\), the controller measures language \(\ell\)'s relative weakness by its validation loss deviation from the multilingual mean. Formally, the pressure score is defined as: 
\[
  \rho_\ell^t =
  \beta \log \frac{L_\ell^t}{\bar L^t}
  + \lambda_s \log s_\ell
  - \lambda_e p_\ell^t ,
\]
The first term amplifies the language-specific loss gap; we set \(\beta=2.0\) to make a moderate 10\% loss difference translate into a meaningful sampling shift, while avoiding overreaction to early-training noise. The second term, \(\lambda_s \log s_\ell\) with \(s_\ell\) as bytes-per-character, is an optional script-complexity prior---we disable it for all final checkpoints (\(\lambda_s=0\)) because our ablations show that this static bias over-corrects Chinese in later training and widens the balance gap. The third term, \(-\lambda_e p_\ell^t\) with \(\lambda_e=0.5\), acts as a negative feedback that gently penalizes languages already receiving high sampling probability, preventing the controller from chasing loss fluctuations too aggressively.

We then convert \(\rho_\ell^t\) into a target distribution \(q_\ell^t\) via softmax with temperature \(\tau\), and smooth the update with momentum \(\eta\):
\[
  q_\ell^t =
  \frac{\exp(\tau\rho_\ell^t)}
       {\sum_k \exp(\tau\rho_k^t)}, \qquad
  \tilde p_\ell^{t+1} = (1-\eta)p_\ell^t + \eta q_\ell^t .
\]
We set \(\tau=2.0\) to produce moderate distributions (avoiding hard one-hot assignments) and \(\eta=0.5\) to balance responsiveness to new signals against stability across updates. Finally, we clip \(\tilde p^{t+1}\) to \([p_{\min}, p_{\max}] = [0.18, 0.50]\) and renormalize. The lower bound ensures that no language drops below 18\% sampling probability (roughly half of uniform 33\%), preventing starvation; the upper bound caps any single language at 50\%, enforcing a balancing constraint that aligns with the 10-epoch BabyLM budget and prevents dominance by one language. The curriculum-dynamic schedule used by submission-facing runs is detailed in Algorithm~\ref{alg:currdyn}.

\section{Experimental Setup}

We first conduct diagnostic experiments to isolate the four challenges identified by ARMOR-DP before training submission-facing checkpoints. These probes use the official gated BabyBabelLM dataset rows available through the Hugging Face dataset viewer and construct equal-length 128-character windows with an equal number of samples for each language. Unless otherwise stated, the focused diagnostic probe uses 180 windows per language, sequence length 64, a 2-layer decoder-only Transformer with hidden size 56 and four attention heads, and is trained for 240 steps over five random seeds. These deliberately short runs isolate mechanisms; they are not used as estimates of full-scale language competence. All benchmark claims come from the 1M- and 100M-budget checkpoints described below. We report the mean validation loss across languages together with the balance gap
\[
\Delta=\max_{\ell}L_{\ell}-\min_{\ell}L_{\ell},
\]
which measures cross-linguistic imbalance. These diagnostic experiments are intended to identify the dominant source of multilingual weakness and guide component selection rather than serve as the final benchmark evaluation.

Based on the diagnostic analysis, we train submission-facing checkpoints by instantiating the three most effective ARMOR-DP components, namely byte-level representation, exposure control, and the dynamic sampling controller. The objective layer is retained as an optional diagnostic module and is evaluated separately through ablation rather than included in the final submission. For byte-level representation, all Unicode text is encoded using standard UTF-8 byte sequences, where each byte is treated as an individual token without language-specific preprocessing or segmentation. Byte values are mapped to training ids by adding an offset of two to reserve ids 0 and 1 for padding and EOS tokens. This mapping is shared throughout training and Hugging Face export, ensuring complete consistency between training and official evaluation.

For the 100M frontier, we train curriculum-dynamic checkpoints with a context length of 512 at model sizes d64, d256, d512, and d768, together with a matched d512 static ZH50 baseline. Training metadata records the actual number of non-padding predicted tokens, wall-clock throughput, per-language consumed epochs, and the maximum consumed epoch, confirming that every reported competition candidate remains below the official ten-epoch limit.

We evaluate all checkpoints using the official BabyLM public zero-shot and fine-tuning pipelines. Before full evaluation, interface compatibility is verified using approximately 100k predicted tokens, 300 official rows per language, training sequence length 128, and exported context length 512. The released equal-trilingual GPT-2 baseline from the official evaluation README is included as a public horizontal reference \citep{babylmeval2026}. Public-suite results are used for model selection because leaderboard-only tasks become available only after submission. The selected d768 checkpoint is finally submitted to the official leaderboard, and the returned multilingual score is reported separately.

Table~\ref{tab:components} maps the checkpoint names used throughout the
results to their active components.

\begin{table*}[t]
\centering
\scriptsize
\begin{tabular}{@{}llllll@{}}
\toprule
System family & Representation & Capacity & Exposure & Controller & Objective \\
\midrule
Balanced adapters & UTF-8 bytes & Adapters & Balanced & None & Token mean \\
Static ZH50 & UTF-8 bytes & Adapters & 25/25/50 & None & Token mean \\
Curriculum-dynamic & UTF-8 bytes & Adapters & Balanced$\rightarrow$adaptive & Loss-based & Token mean \\
Dynamic + pressure & UTF-8 bytes & Adapters & Adaptive & Loss-based & Pressure-aware \\
Curr.-dyn. + pressure & UTF-8 bytes & Adapters & Balanced$\rightarrow$adaptive & Loss-based & Pressure-aware \\
\bottomrule
\end{tabular}
\caption{Component map for the checkpoint families reported in
Sections~4--7. The d256, d512, and d768 final systems are capacity-scaled
instances of curriculum-dynamic.}
\label{tab:components}
\end{table*}

\section{Results}

\subsection{Balanced input is not balanced pressure}

Figure~\ref{fig:overview} compresses the first failure mode behind ARMOR-DP:
a balanced multilingual budget is not automatically a balanced learning
pressure. With the same 23,040-character budget per language, Chinese contains
66,228 UTF-8 bytes and 1,535 unique characters, while English and Dutch each
have about 23,000 bytes and fewer than 60 unique characters.

This pressure is visible in Table~\ref{tab:armor}. A character-level balanced
baseline collapses on Chinese, with a mean loss of 3.066 and a balance gap of
1.858. Switching only the representation to byte units reduces mean loss to
2.362 and gap to 0.185. This is the clearest evidence for the first
ARMOR-DP layer.

\begin{table*}[t]
\centering
\small
\begin{tabular}{lrrrrr}
\toprule
Condition & Mean loss & Gap & EN & NL & ZH \\
\midrule
Char balanced baseline & 3.066 $\pm$ 0.020 & 1.858 $\pm$ 0.051 & 2.509 & 2.415 & 4.273 \\
Byte balanced baseline & 2.362 $\pm$ 0.007 & 0.185 $\pm$ 0.063 & 2.455 & 2.361 & 2.269 \\
Byte + ZH50 baseline & 2.317 $\pm$ 0.012 & 0.548 $\pm$ 0.042 & 2.531 & 2.435 & 1.983 \\
Byte + ZH50 adapters & 2.235 $\pm$ 0.007 & 0.511 $\pm$ 0.028 & 2.436 & 2.344 & 1.926 \\
Byte balanced$\rightarrow$ZH50 adapters & 2.246 $\pm$ 0.010 & 0.408 $\pm$ 0.022 & 2.417 & 2.312 & 2.008 \\
Byte + ZH50 routing gate & 2.272 $\pm$ 0.014 & 0.459 $\pm$ 0.051 & 2.457 & 2.361 & 1.998 \\
Byte + ZH50 adapters + rare loss & 2.276 $\pm$ 0.007 & 0.663 $\pm$ 0.032 & 2.532 & 2.425 & 1.869 \\
\bottomrule
\end{tabular}
\caption{Controlled ARMOR-DP probe on official BabyBabelLM rows. Representation
armor removes the Chinese collapse; exposure and capacity improve mean loss
but can transfer worst-language status to English.}
\label{tab:armor}
\end{table*}

\begin{figure}[t]
  \centering
  \makebox[\linewidth][c]{\includegraphics[width=1\linewidth]{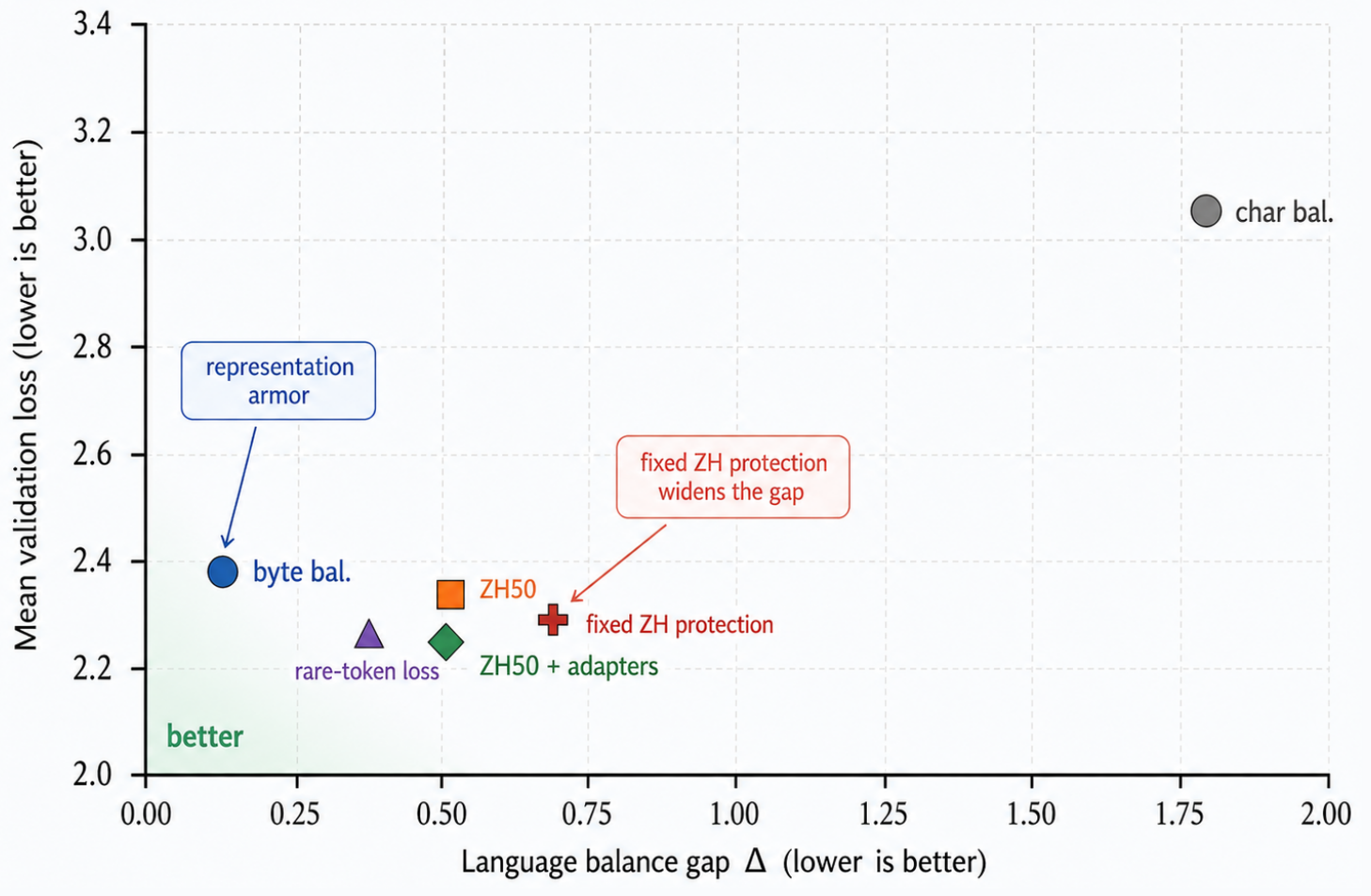}}
  \caption{End-of-run score/balance tradeoff across diagnostic systems. The
  axes are final mean validation loss and final language gap (not training
  steps); lower is better on both axes. Byte representation removes
  the first collapse, while ZH-protective adapters improve mean loss but widen
  the language gap, making English the weakest language.}
  \label{fig:tradeoff}
\end{figure}

\subsection{Protection moves the bottleneck}

Exposure and capacity armor improve the average score but change which language
needs protection. Figure~\ref{fig:tradeoff} makes this tradeoff visible without
requiring the reader to scan every loss column in Table~\ref{tab:armor}. Byte +
ZH50 adapters obtains the lowest mean loss in the controlled probe, but the
worst language is English in all seeds. This transfer is not a failure of the
framework; it is the phenomenon the framework should explain. A multilingual
BabyLM system should not protect a fixed language. It should protect the
language currently under the highest developmental pressure.

\subsection{Dynamic control turns movement into signal}

We therefore ran a dynamic probe. Every 24 steps, the controller evaluates
per-language losses and updates sampling ratios with capped adaptive protection.
The pure dynamic adapter system keeps the balance gap close to balanced
training while slightly lowering mean loss. A script-prior variant pushes more
mass to Chinese and achieves lower mean loss than pure dynamic control, but at a
larger gap. Static ZH50 remains the lowest mean-loss configuration in this
small run, but has the largest gap among the adapter systems.

The logged update points also provide a direct trace of weakest-language
movement. In the objective probe, the weakest language often begins as Dutch,
moves to Chinese during early byte-level learning, and then transfers to
English after Chinese-protective exposure has taken effect. Across three seeds,
the adaptive exposure controller shows two NL$\rightarrow$ZH transitions and
three ZH$\rightarrow$EN transitions. This supports the main claim that
``weakest language'' is a training-time state, not a fixed label.

\begin{table}[t]
\centering
\small
\begin{tabular}{lrr}
\toprule
Condition & Mean loss & Gap \\
\midrule
Balanced adapters & 2.230 & 0.179 \\
Dynamic-loss adapters & 2.229 & 0.199 \\
Dynamic-loss routing gate & 2.264 & 0.168 \\
Dynamic + script prior adapters & 2.217 & 0.290 \\
Curriculum adapters & 2.216 & 0.353 \\
Static ZH50 adapters & 2.198 & 0.489 \\
\bottomrule
\end{tabular}
\caption{Dynamic protection probe over three seeds. Dynamic loss control is
most useful as a balance-preserving controller; static ZH50 still optimizes
mean loss at the cost of English.}
\label{tab:dynamic}
\end{table}

\begin{figure}[t]
  \centering
  \makebox[\linewidth][c]{\includegraphics[width=1\linewidth]{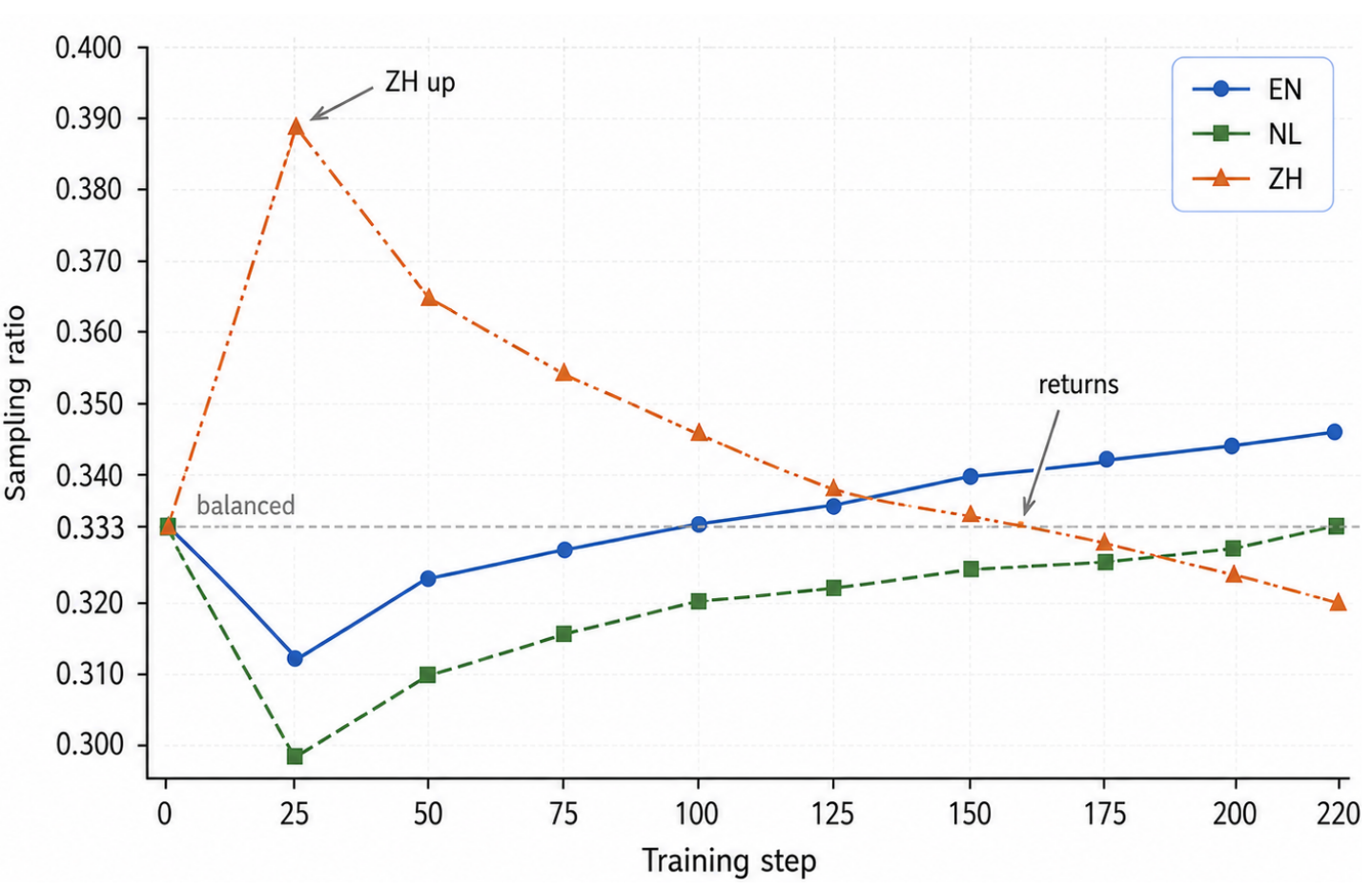}}
  \caption{How the dynamic controller changes exposure. Each curve is the
  sampling probability for one language, and the x-axis is the logged
  controller-update index from a dynamic probe (unlike the end-of-run axes in
  Figure~\ref{fig:tradeoff}). The controller
  first raises Chinese exposure, then returns probability mass to English and
  Dutch as Chinese becomes easier, showing why fixed ZH protection can become
  stale.}
  \label{fig:dynamic}
\end{figure}

\begin{figure}[t]
  \centering
  \makebox[\linewidth][c]{\includegraphics[width=1\linewidth]{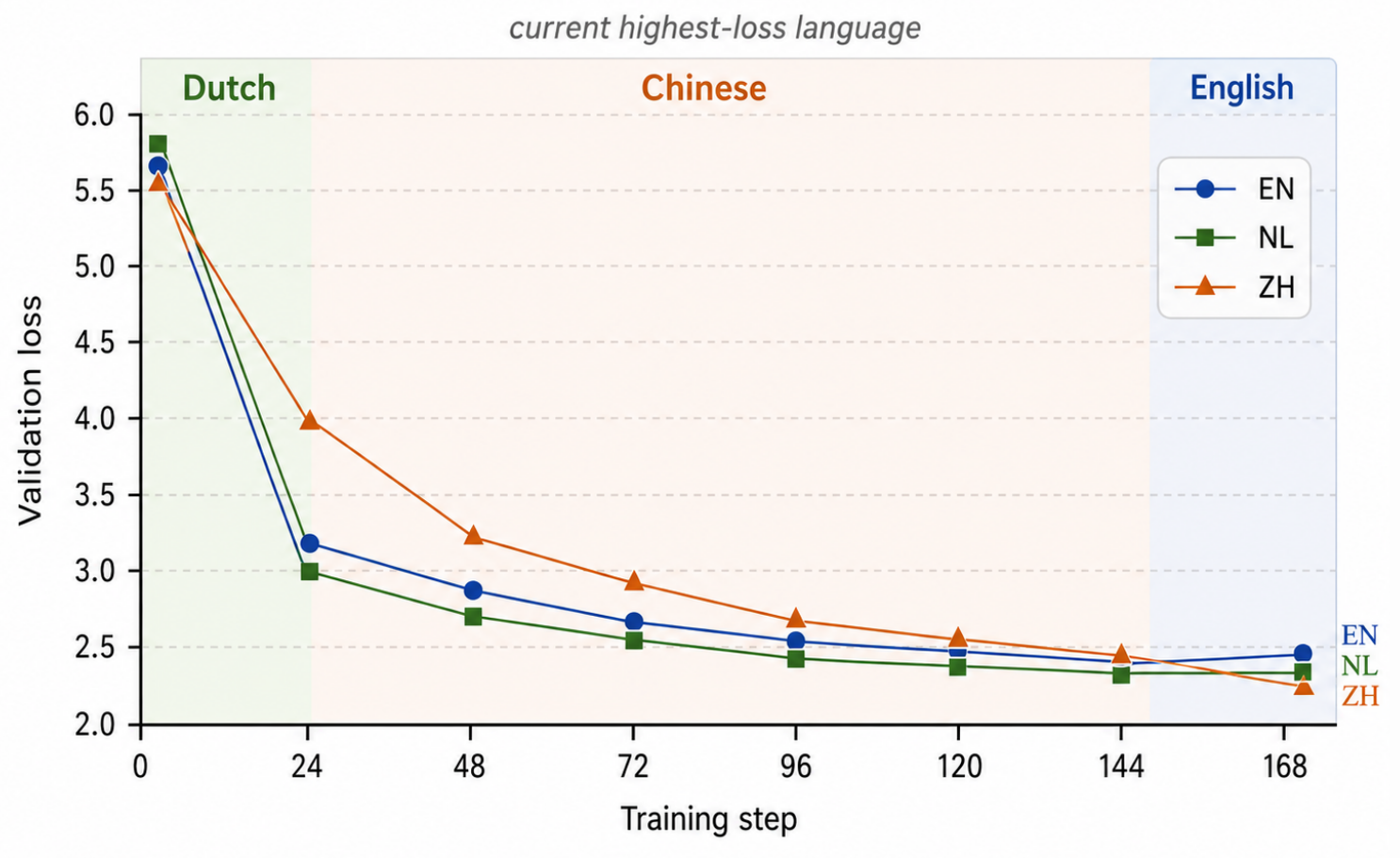}}
  \caption{Weakest-language identity over training. The highlighted language at
  each step is the current highest-loss language. The bottleneck moves from
  Dutch to Chinese and then to English, supporting the claim that weakness is a
  training-time state rather than a fixed language label.}
  \label{fig:transition}
\end{figure}

\subsection{Objectives must not repeat exposure}

Before comparing objective variants, we distinguish genuine pressure weighting
from a reduction artifact: averaging losses by language already equalizes
languages even when explicit pressure weights are weak.

Table~\ref{tab:objective} compares objective variants under byte-level
representation and language adapters. Naive rare-token weighting is the least
attractive objective in this confirmation run: it lowers Chinese loss but
widens the final gap by making English much worse. Pressure-aware weighting is
more controlled. Under a fixed ZH50 exposure schedule, it reduces the gap from
0.348 to 0.232 with only a modest mean-loss cost. However, it still does not
beat the balance-preserving adaptive exposure baseline. We therefore treat
objective armor as necessary but not solved: the evidence favors a
pressure-aware objective, while the final system should tune its strength so it
does not duplicate exposure protection.

We later found that objective design has one additional pitfall. A
language-level pressure loss can improve balance simply by changing the
reduction unit from token means to language means, even when pressure weights
are weak. To separate the objective layer from this hidden equalization effect,
we added a token-level pressure reduction with warmup, end decay, and a capped
maximum weight of 1.4. With strength 0.20, this scheduled objective reduces
the local 1M validation gap from 0.253 to 0.240 while keeping mean loss
essentially unchanged (2.264 to 2.262). On the public zero-shot suite, it
reduces the curriculum-dynamic language gap from 0.0359 to 0.0330, but lowers
overall accuracy from 0.4634 to 0.4612. We therefore keep curriculum-dynamic as
the scorer and treat scheduled token pressure as evidence for the objective
layer rather than as the final checkpoint.

\begin{table}[t]
\centering
\small
\begin{tabular}{@{}lrrr@{}}
\toprule
Condition & Mean loss & Gap & Worst \\
\midrule
Adaptive exp. & 2.344 & 0.118 & EN \\
Balanced & 2.346 & 0.119 & EN \\
Adaptive exp. + pressure & 2.360 & 0.158 & EN \\
ZH50 + pressure & 2.354 & 0.232 & EN \\
ZH50 & 2.320 & 0.348 & EN \\
ZH50 + rare token & 2.359 & 0.481 & EN \\
\bottomrule
\end{tabular}
\caption{Objective-layer probe over three seeds. Pressure-aware weighting is
substantially less damaging than rare-token weighting and reduces the
over-protection gap under ZH50, but adaptive exposure remains the better
balance-preserving controller.}
\label{tab:objective}
\end{table}

\subsection{Public evaluation separates score and diagnosis}

Table~\ref{tab:officialzero} reports a full pass through the official
multilingual zero-shot task groups for the original five 1M-token checkpoints
plus the best scheduled token-pressure variant. These
are not interface smoke tests: each checkpoint is loaded through
the HuggingFace causal-LM path and evaluated with the official English, Dutch,
and Chinese \texttt{lm-eval} task definitions. Hidden leaderboard tasks are not
available locally, so the table should be read as public-suite evidence.

The official score changes the model-selection story. Static ZH50 remains the
best local language-modeling-loss checkpoint, but it is not the strongest
zero-shot system. Curriculum-dynamic protection obtains the best mean score by
a small margin over balanced byte adapters, and dynamic+pressure obtains the
smallest language gap and variance while sacrificing mean accuracy. This is
the desired ARMOR-DP separation: the competition-facing scorer, the
balance-preserving controller, and the diagnostic pressure objective need not
be the same checkpoint.

\begin{table*}[t]
\centering
\small
\begin{tabular}{lrrrrrrr}
\toprule
System & Overall & Worst & Gap & Var. & EN & NL & ZH \\
\midrule
Balanced adapters & 0.4633 & NL 0.4465 & 0.0334 & 0.00019 & 0.4798 & 0.4465 & 0.4635 \\
Static ZH50 & 0.4574 & NL 0.4413 & 0.0379 & 0.00025 & 0.4791 & 0.4413 & 0.4519 \\
Curriculum-dynamic & \textbf{0.4634} & NL 0.4439 & 0.0359 & 0.00022 & 0.4798 & 0.4439 & \textbf{0.4665} \\
Curr.-dyn. + token pressure & 0.4612 & NL 0.4443 & 0.0330 & 0.00018 & 0.4772 & 0.4443 & 0.4620 \\
Dynamic + pressure & 0.4541 & NL 0.4443 & \textbf{0.0181} & \textbf{0.00006} & 0.4623 & 0.4443 & 0.4558 \\
Curriculum-dynamic + pressure & 0.4554 & NL 0.4438 & 0.0292 & 0.00016 & 0.4730 & 0.4438 & 0.4495 \\
\bottomrule
\end{tabular}
\caption{Full public official zero-shot evaluation for 1M-token checkpoints.
Overall is the mean over English, Dutch, and Chinese task-group means. Gap and
variance are computed over language means, so lower is more balanced.}
\label{tab:officialzero}
\end{table*}

We evaluate all checkpoints using the official multilingual fine-tuning pipeline over 22 public task-language pairs. Since fine-tuning losses are not directly comparable across task-specific classifiers, Table~\ref{tab:officialfine} reports the official accuracy metric. Curriculum-dynamic achieves the highest fine-tuning mean and the best average of zero-shot and fine-tuning performance, primarily driven by improved Chinese accuracy, which increases from 0.210 under balanced adapters and 0.207 under static ZH50 to 0.244. Pressure-aware variants instead prioritize cross-lingual balance, with curriculum-dynamic+pressure achieving the smallest fine-tuning gap while shifting the weakest language from Chinese to Dutch.

\begin{table*}[t]
\centering
\small
\begin{tabular}{lrrrrrrrr}
\toprule
System & Fine & Combined & Worst & Gap & Var. & EN & NL & ZH \\
\midrule
Balanced adapters & 0.2326 & 0.3479 & ZH 0.2099 & 0.0485 & 0.00040 & 0.2584 & 0.2295 & 0.2099 \\
Static ZH50 & 0.2254 & 0.3414 & ZH 0.2072 & 0.0391 & 0.00026 & 0.2464 & 0.2226 & 0.2072 \\
Curriculum-dynamic & \textbf{0.2395} & \textbf{0.3514} & NL 0.2180 & 0.0388 & 0.00026 & 0.2567 & 0.2180 & \textbf{0.2437} \\
Dynamic + pressure & 0.2228 & 0.3385 & ZH 0.2094 & 0.0337 & 0.00021 & 0.2159 & \textbf{0.2431} & 0.2094 \\
Curriculum-dynamic + pressure & 0.2364 & 0.3459 & NL 0.2304 & \textbf{0.0143} & \textbf{0.00004} & 0.2447 & 0.2304 & 0.2342 \\
\bottomrule
\end{tabular}
\caption{Full public official fine-tuning evaluation. Fine is the mean over
language-level task means. Combined is the unweighted average of the zero-shot
overall score in Table~\ref{tab:officialzero} and the fine-tuning mean here.
Gap and variance are computed over fine-tuning language means.}
\label{tab:officialfine}
\end{table*}

\subsection{Focused 1M grid}

The next local experiment scales the most relevant systems from the 100k-token
interface probe to 1M predicted tokens, using 600 cached official rows per
language. Table~\ref{tab:focusedgrid} shows the 1M local validation losses
and the actual consumed epochs logged by the training sampler. All runs remain
below the 10-epoch competition limit. Static ZH50 obtains the lowest mean loss
but does so by driving Chinese loss far below English and Dutch, making
English the weakest language and widening the gap to 0.389. Curriculum-dynamic
protection gives up 0.015 mean loss while reducing the gap to 0.253. Balanced
byte adapters are also highly competitive, which reinforces the representation
claim: script-fair representation is a prerequisite before exposure control can
be interpreted.

Although dynamic+pressure and curriculum-dynamic+pressure improve cross-lingual balance, they incur a reduction in the overall mean score compared with curriculum-dynamic alone. We therefore retain pressure-aware weighting as an optional fourth-layer mechanism for diagnostic analysis rather than including it in the final submission, suggesting that its effectiveness depends on appropriate strength scheduling.

\begin{table*}[t]
\centering
\small
\begin{tabular}{lrrrrrrr}
\toprule
System & Mean & Gap & Worst & EN loss & NL loss & ZH loss & Max ep. \\
\midrule
Balanced adapters & 2.258 & 0.226 & EN & 2.372 & 2.258 & 2.145 & 4.410 \\
Static ZH50 & \textbf{2.248} & 0.389 & EN & 2.414 & 2.305 & 2.025 & 6.555 \\
Curriculum-dynamic & 2.264 & 0.253 & EN & 2.389 & 2.266 & 2.136 & 4.523 \\
Dynamic + pressure & 2.295 & 0.337 & EN & 2.438 & 2.345 & 2.101 & 4.387 \\
Curriculum-dynamic + pressure & 2.279 & 0.273 & EN & 2.407 & 2.298 & 2.133 & 4.505 \\
\bottomrule
\end{tabular}
\caption{Focused 1M-token grid on cached official rows. The epoch column is
the maximum actual consumed epoch over English, Dutch, and Chinese, recorded
in each checkpoint's training metadata.}
\label{tab:focusedgrid}
\end{table*}

\subsection{Capacity scaling makes dynamic protection competitive}

The official public suite requires an interface that can tolerate long
zero-shot contexts. We therefore closed a 512-context line after the 1M public
evaluation. A first budget curve with the tiny d64 model showed that simply
training longer was not enough: the best 100M d64 curriculum-dynamic run
reached only 0.333 combined. A substantial improvement came from capacity
armor. Increasing the same curriculum-dynamic system to d256, d512, and d768
raises the public combined score to 0.371, 0.381, and 0.387 respectively.

\begin{figure}[t]
\centering
\makebox[\linewidth][c]{\includegraphics[width=1\linewidth]{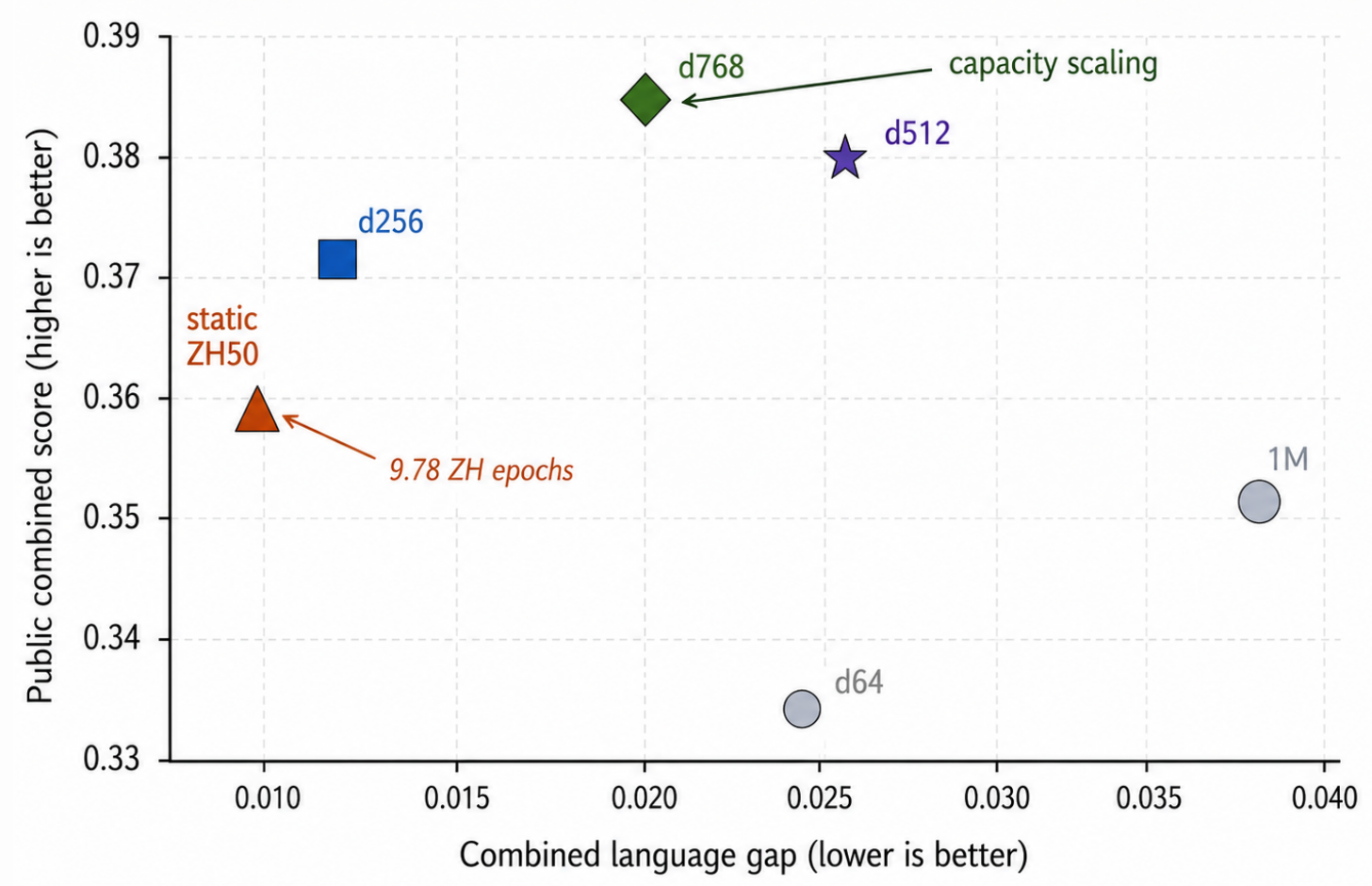}}
\caption{Internal ARMOR-DP public score/balance frontier for the 100M line.
The released GPT-2 equal-trilingual baseline is higher (0.4097 combined) and is
reported in Table~\ref{tab:scaling}, but is not drawn because it is an external
organizer baseline rather than an ARMOR-DP variant. Within the internal
frontier, capacity scaling turns curriculum-dynamic protection into the
best-performing ARMOR-DP scorer, while the matched static ZH50 baseline sits lower
despite a small public gap.}
\label{fig:scalingfrontier}
\end{figure}

This changes the ARMOR-DP submission decision. The 100M d768 checkpoint is the
strongest ARMOR-DP public scorer and also has a competitive public language
gap, while the d512 checkpoint provides the clearest matched-capacity
comparison against static ZH50. A matched d512 static ZH50 baseline
falls to 0.359 combined, and its local language-modeling gap is much larger:
0.495 versus 0.079 for d512 curriculum-dynamic. Static ZH50 also consumes 9.78
Chinese epochs, close to the 10-epoch competition limit, whereas
curriculum-dynamic remains below 6.61 epochs for every language. Thus
ARMOR-DP's dynamic exposure is no longer only diagnostic at the final scale; it
is also the stronger public-suite system under matched capacity. The released
equal-trilingual GPT-2 baseline still sets a higher public combined reference
(0.410 versus 0.387), mainly through zero-shot accuracy, while d768 is
comparable on fine-tuning. This makes the current award-facing claim sharper:
ARMOR-DP explains and improves the dynamic-protection frontier, but it should
not be described as a finished leaderboard winner before hidden evaluation.
The organizer GPT-2 uses a conventional subword tokenizer, whereas ARMOR-DP
uses bytes; a token therefore does not represent the same amount of linguistic
content in the two systems. We include GPT-2 only as an external score
reference, not as a controlled tokenizer comparison.
Figure~\ref{fig:scalingfrontier} turns this model-selection result into the
frontier view used for the final submission decision: moving from d64 to larger
curriculum-dynamic models improves score without returning to the large local
gap of static ZH50.

\begin{table*}[t]
\centering
\small
\begin{tabular}{lrrrrrr}
\toprule
System & Non-pad tok. & Zero & Fine & Comb. & Worst comb. & Gap \\
\midrule
1M curr.-dyn. & 1.0M & 0.4634 & 0.2395 & 0.3514 & NL 0.3309 & 0.0373 \\
100M d64 curr.-dyn. & 87.3M & 0.4692 & 0.1976 & 0.3334 & NL 0.3202 & 0.0248 \\
100M d256 curr.-dyn. & 87.3M & 0.4842 & 0.2587 & 0.3714 & NL 0.3663 & \textbf{0.0118} \\
100M d512 curr.-dyn. & 87.0M & 0.4954 & 0.2659 & 0.3807 & NL 0.3667 & 0.0256 \\
100M d768 curr.-dyn. & 87.0M & 0.5032 & \textbf{0.2699} & 0.3866 & NL 0.3734 & 0.0200 \\
100M d512 static ZH50 & 89.0M & 0.4845 & 0.2330 & 0.3587 & EN 0.3538 & 0.0095 \\
Official GPT-2 tri. baseline & -- & \textbf{0.5516} & 0.2678 & \textbf{0.4097} & ZH \textbf{0.3929} & 0.0306 \\
\bottomrule
\end{tabular}
\caption{Public official capacity frontier. ``Non-pad tok.`` counts actual
predicted non-padding tokens; all ARMOR-DP rows remain under the 10-epoch
competition cap. The d512 curr.-dyn. and d512 static rows are
matched-capacity systems. The GPT-2 row is the released equal-trilingual
baseline from the official evaluation README; its subword tokens are not
content-equivalent to ARMOR-DP's byte tokens.}
\label{tab:scaling}
\end{table*}

\paragraph{Final leaderboard submission.}
We submit the d768 curriculum-dynamic checkpoint as the final challenge-facing
ARMOR-DP system. The official leaderboard accepts the submission as challenge
eligible and returns a multilingual average of 38.32. The language averages are
35.34 for English, 40.96 for Dutch, and 38.65 for Chinese. This result should be
read as a final submission check rather than a new model-selection signal: all
system choices above were made from public official evaluations, and the
leaderboard result confirms the submitted checkpoint while preserving the paper's
main claim as diagnostic rather than leaderboard dominance.

\section{Related Work}

ARMOR-DP builds on three lines of work. First, BabyLM studies sample-efficient
pretraining under developmentally plausible data budgets
\citep{babylm2026guidelines,babybabellm2026,babylmfindings2025}. Our setting
adds the multilingual constraint that a fixed token or character budget can
interact with script and morphology. Second, token-free and byte-level models
show that character and byte units can reduce tokenizer-specific assumptions
\citep{xue2022byt5,clark2022canine,tay2022charformer}; here we use bytes as a
diagnostic representation, not as a claim that bytes are universally optimal.
Third, multilingual sampling, adapters, curriculum learning, sparse routing,
and loss reweighting provide tools for allocating exposure, capacity, and
gradient pressure \citep{conneau2020xlmr,houlsby2019adapters,pfeiffer2020madx,bengio2009curriculum,fedus2022switch,lin2017focal}.
ARMOR-DP's contribution is to combine these tools around the moving identity
of the weakest language.

\section{Discussion}

Our experiments show that multilingual imbalance is an evolving training state:
the dominant weakness shifts from representation to exposure and then capacity,
so static language protection becomes progressively less effective. ARMOR-DP
separates these factors and their interventions, allowing adaptation to follow
the current learning state rather than a predetermined schedule.

For the BabyLM challenge, byte-level representation, exposure control, and the
dynamic controller consistently improve multilingual performance and form our
final system. The objective layer is optional: pressure-aware weighting reduces
cross-lingual gaps but modestly lowers the mean score, suggesting that its
strength should be scheduled rather than applied unconditionally. This modular
separation also makes interactions and future combinations easier to analyze.

\section{Submission Choice}

The final submission is the curriculum-dynamic d768 checkpoint, which has the
best overall public performance while maintaining competitive cross-lingual
balance. We retain the matched d512 checkpoint as the principal diagnostic
control against static ZH50.

\section{Conclusion}

ARMOR-DP reframes multilingual BabyLM as protection of the language currently
under greatest developmental pressure. Across diagnostics, 1M probes, and
capacity scaling, representation establishes a script-fair foundation, exposure
and capacity improve performance, and dynamic control tracks bottleneck
transitions that fixed protection misses. By separating diagnosis from modular,
empirically instantiable interventions, ARMOR-DP offers a flexible basis for
future multilingual training.

\section*{Limitations}

The experiments use small diagnostic models and cached subsets of official dataset
rows. The public official suite is complete for the checkpoints reported here,
and we include the released GPT-2 equal-trilingual public baseline for
orientation. Leaderboard-only tasks were not available during model selection;
they were scored only after the final d768 checkpoint was submitted. We
therefore treat the official leaderboard average as a post-selection submission
result, not as a tuning signal. The 100M-budget runs use a
100M maximum-token training budget, but actual predicted non-padding tokens
are lower because many cached examples are shorter than 512 tokens. The 100M
capacity frontier is single-seed because of compute cost; diagnostic claims are
supported by multi-seed small probes, but large-scale score variance remains
unmeasured. The controller is evaluated on only three languages, and a larger
language set may produce noisier pressure signals and require different update
bounds. We also report one controller configuration rather than a full
sensitivity sweep. The script-complexity prior was the most brittle component:
it over-corrected Chinese later in training and was disabled in final runs. A
scheduled prior that decays after early training remains an open direction.

\bibliography{references}

\clearpage
\appendix
\onecolumn
\captionsetup{hypcap=false}

\section{Reproducibility Details}

The main paper keeps the eight-page submission concise. This appendix records
the implementation details that are most likely to matter for reproducing or
auditing the ARMOR-DP claim.

\paragraph{Data and tokenization.}
All controlled diagnostics use English, Dutch, and Chinese rows from the
BabyBabelLM public data interface. We form equal-length windows per language
for the diagnostic probes so that representation and exposure interventions can
be compared under matched surface budgets. Submission-facing checkpoints use a
UTF-8 byte tokenizer with the same id mapping in training and HuggingFace
export. Padding labels are masked from the language-modeling loss, and all
reported token counts are predicted non-padding tokens rather than padded
sequence slots.

\paragraph{Evaluation.}
The public official suite is run through the HuggingFace causal-LM loading path
for zero-shot tasks and through the official multilingual fine-tuning pipeline
for supervised tasks. Hidden leaderboard tasks are not locally available. The
fine-tuning pass contains 22 public task-language pairs for the 100M capacity
frontier, with no missing tasks.

\paragraph{Epoch accounting.}
The training loop records consumed epochs per language by counting the actual
examples sampled for that language. This is stricter than reporting optimizer
steps alone: a static ZH50 mixer can keep the global token budget modest while
nearly exhausting the Chinese-language epoch cap. All competition-facing rows
reported in the main paper and in the ledger below remain below the 10-epoch
limit.

\begin{center}
\footnotesize
\begin{tabular}{@{}lrrrrrrrrr@{}}
\toprule
System & Tok. (M) & Seq. & Batch & d/layers & Steps & EN ep. & NL ep. & ZH ep. & Wall min \\
\midrule
1M balanced adapters & 0.994 & 128 & 8 & 64/2 & 985 & 4.32 & 4.40 & 4.41 & -- \\
1M static ZH50 & 0.996 & 128 & 8 & 64/2 & 985 & 3.26 & 3.32 & 6.56 & -- \\
1M curr.-dyn. & 0.994 & 128 & 8 & 64/2 & 985 & 4.34 & 4.28 & 4.52 & -- \\
1M dyn.+pressure & 0.994 & 128 & 8 & 64/2 & 985 & 4.38 & 4.37 & 4.39 & -- \\
1M curr.-dyn.+pressure & 0.994 & 128 & 8 & 64/2 & 985 & 4.34 & 4.29 & 4.51 & -- \\
5M l512 curr.-dyn. & 4.402 & 512 & 2 & 64/2 & 4,893 & 1.58 & 1.59 & 1.72 & -- \\
5M l512 static ZH50 & 4.520 & 512 & 8 & 64/2 & 1,224 & 1.22 & 1.19 & 2.49 & -- \\
100M d256 curr.-dyn. & 87.272 & 512 & 8 & 256/6 & 24,462 & 6.46 & 6.35 & 6.76 & 26.8 \\
100M d512 curr.-dyn. & 87.037 & 512 & 8 & 512/8 & 24,462 & 6.56 & 6.54 & 6.46 & 38.7 \\
100M d768 curr.-dyn. & 86.992 & 512 & 4 & 768/12 & 48,924 & 6.61 & 6.59 & 6.37 & 108.7 \\
100M d512 static ZH50 & 89.011 & 512 & 8 & 512/8 & 24,462 & 4.91 & 4.87 & 9.78 & 32.7 \\
\bottomrule
\end{tabular}
\captionof{table}{Training-accounting ledger. Epochs are actual consumed epochs for
English, Dutch, and Chinese, not estimates from optimizer steps. ``Wall min''
is reported when available from the remote training metadata.}
\label{tab:epochaccount}
\end{center}

\section{Hyperparameters and Controller Settings}

\begin{center}
\scriptsize
\begin{tabular}{@{}lll@{}}
\toprule
Component & Setting & Value \\
\midrule
Diagnostic corpus & Windows/language & 180 for dynamic and objective probes \\
Diagnostic model & Architecture & Decoder-only Transformer, d=56, 2 layers, 4 heads \\
Diagnostic training & Steps/seeds & 240 steps for dynamic probe; 180 steps for objective probe; three seeds \\
Rare-token control & Token weighting & Corpus-frequency derived; rarer bytes weighted more; clipped to 0.25--1.8 \\
1M public checkpoints & Architecture & d=64, 2 layers, 4 heads, language adapters \\
1M public checkpoints & Training/export context & 128-token training context; 512-token exported context \\
100M capacity frontier & Architecture & d256/6 layers, d512/8 layers, d768/12 layers \\
100M capacity frontier & Optimizer learning rate & 3e-4 for d256/d512; 2e-4 for d768 \\
Static ZH50 & Sampling ratio & EN/NL/ZH = 0.25/0.25/0.50 \\
Curriculum-dynamic & Start schedule & Balanced sampling, then Chinese-protective/adaptive sampling \\
Adaptive controller & Update interval & 24 diagnostic; 80 for 1M/5M/d768; 40 for d256/d512 \\
Adaptive controller & Bounds & \(p_{\min}=0.18,\ p_{\max}=0.50\) \\
Adaptive controller & Loss/smoothing/temp. & \(\beta=2.0,\ \eta=0.5,\ \tau=2.0\) \\
Adaptive controller & Exposure penalty & \(\lambda_e=0.5\) for curriculum-dynamic runs \\
Pressure objective & Weight bounds & 0.25--1.8 in diagnostic runs; capped at 1.4 in tuned token-pressure probes \\
\bottomrule
\end{tabular}
\captionof{table}{Core hyperparameters. All competition-facing runs set
\texttt{max\_epochs=10}; the actual consumed-epoch ledger is in
Table~\ref{tab:epochaccount}.}
\label{tab:hyperparams}
\end{center}

\paragraph{Controller pseudocode.}
\algcaption{Curriculum-dynamic exposure controller.}\label{alg:currdyn}
Initialize \(p=(1/3,1/3,1/3)\); train balanced until \(T/2\); set
\(p=(0.25,0.25,0.50)\); after \(T/2\), every \(K\) steps, evaluate
\(L_\ell^t\), compute \(\rho_\ell^t\) and \(q_\ell^t\) with
\(\beta=2.0,\eta=0.5,\tau=2.0,\lambda_e=0.5,\lambda_s=0\), then clip to
\([0.18,0.50]\), renormalize, and continue.

\paragraph{Controller sensitivity.}
We did not run an exhaustive sensitivity sweep. The clearest sensitivity was to
the script prior: enabling \(\lambda_s>0\) could help Chinese early but
over-corrected it later, so all final checkpoints use \(\lambda_s=0\). A
scheduled, decaying script prior may preserve the early benefit without the
late bias. The behavior of the bounds and update rule with substantially more
than three languages is left for future work.

\section{Additional Public Evaluation Details}

Table~\ref{tab:fiveandofficial} adds the 5M 512-context public suite that is
summarized only implicitly in the main paper. At 5M, curriculum-dynamic already
beats static ZH50 on combined score, mostly because static ZH50 loses more
fine-tuning accuracy. At 100M, the same direction becomes the submission-facing
line.

\begin{center}
\footnotesize
\begin{tabular}{@{}lrrrrrrrr@{}}
\toprule
System & Zero & Fine & Comb. & EN comb. & NL comb. & ZH comb. & Worst & Gap \\
\midrule
5M l512 curr.-dyn. & 0.464 & 0.216 & 0.340 & 0.360 & 0.332 & 0.329 & ZH 0.329 & 0.031 \\
5M l512 static ZH50 & 0.466 & 0.189 & 0.328 & 0.338 & 0.307 & 0.338 & NL 0.307 & 0.031 \\
100M d256 curr.-dyn. & 0.484 & 0.259 & 0.371 & 0.370 & 0.366 & 0.378 & NL 0.366 & 0.012 \\
100M d512 curr.-dyn. & 0.495 & 0.266 & 0.381 & 0.392 & 0.367 & 0.383 & NL 0.367 & 0.026 \\
100M d768 curr.-dyn. & 0.503 & 0.270 & 0.387 & 0.393 & 0.373 & 0.393 & NL 0.373 & 0.020 \\
100M d512 static ZH50 & 0.485 & 0.233 & 0.359 & 0.354 & 0.363 & 0.359 & EN 0.354 & 0.009 \\
\bottomrule
\end{tabular}
\captionof{table}{Additional public official combined results. Combined is the
unweighted average of zero-shot and fine-tuning means.}
\label{tab:fiveandofficial}
\end{center}

Table~\ref{tab:d768tasks} expands the public fine-tuning tasks for the final
100M d768 curriculum-dynamic scorer. The table is not used to tune claims in
the main paper; it is included to make clear which public task groups underlie
the reported fine-tuning mean.

\begin{center}
\footnotesize
\begin{tabular}{@{}lrrr@{}}
\toprule
Task & EN & NL & ZH \\
\midrule
ARC & 0.235 & 0.235 & 0.262 \\
Belebele & 0.193 & 0.233 & 0.165 \\
BMLAMA & 0.123 & 0.113 & 0.130 \\
MNLI & 0.436 & 0.365 & 0.408 \\
SIB200 & 0.150 & 0.305 & 0.425 \\
TruthfulQA & 0.232 & 0.268 & 0.214 \\
XNLI & 0.436 & -- & 0.427 \\
INCLUDE & -- & 0.321 & 0.277 \\
\bottomrule
\end{tabular}
\captionof{table}{Public fine-tuning task accuracies for the 100M d768
curriculum-dynamic checkpoint. Dashes indicate language-task pairs not present
in the local public suite.}
\label{tab:d768tasks}
\end{center}

\section{Negative Results and Deferred Directions}

We preserve the failed or deferred branches because they help distinguish the
framework from a post-hoc winning recipe. Static ZH50 is retained as the
matched baseline that makes the dynamic comparison meaningful, rather than
discarded after the final system is chosen.

\paragraph{Simple gates were not enough.}
The original gate direction was too weak when treated as a generic attention
modification. A language-conditioned routing gate became more interpretable as
capacity armor and obtained a reasonable diagnostic tradeoff, but it did not
beat adapters as the main scorer. We therefore use adapters in the final
competition-facing line and keep routing gates as a capacity diagnostic.

\paragraph{Rare-token weighting over-protects Chinese.}
Rare-token weighting lowers Chinese loss after Chinese-protective exposure has
already helped, but it widens the balance gap by making English the weakest
language. In the objective probe, ZH50+rare reaches a gap of 0.481, compared
with 0.348 for ZH50 and 0.118 for adaptive exposure. We treat this as a
negative result for frequency-only objectives.

\paragraph{Pressure-aware weighting is useful but not final.}
Pressure-aware weighting aligns better with ARMOR-DP than rare-token weighting
and gives the smallest public zero-shot gap in the 1M suite and the smallest
public fine-tuning gap with curriculum-dynamic exposure. Because it lowers mean
public score in the current tuning range, the final scorer omits it; pressure
objectives remain a fourth-layer research direction.

\end{document}